\documentclass{IEEEtran}

 \usepackage{amsmath,amsthm,amssymb,amsfonts}
\usepackage{hyperref}
\usepackage{amssymb}
\usepackage{array,multirow,graphicx}
\usepackage{float}
\usepackage{balance}
\usepackage{subfigure}
\usepackage{verbatim}

\title{Efficient Motion Modelling with Variable-sized blocks from Hierarchical Cuboidal Partitioning}

 \author{\IEEEauthorblockN{Priyabrata Karmakar, Manzur Murshed, Manoranjan Paul, David Taubman} 
	
	\thanks{Priyabrata Karmakar and Manzur Murshed are with the Institute of Innovation, Science and Sustainability, Federation University Australia (e-mail:\{p.karmakar, manzur.murshed\}@federation.edu.au).
		
	Manoranjan Paul is with the School of Computing
	and Mathematics, Charles Sturt University, Australia (email: mpaul@csu.edu.au).
	
	David Taubman is with the School of Electrical Engineering and
	Telecommunications, University of New South Wales, Australia (email: d.taubman@unsw.edu.au)
		
		}
}

\begin{document}
	
%\begin{frontmatter}

%\title{Efficient Motion Modelling with Variable-sized blocks from Hierarchical Cuboidal Partitioning}
%
% Single address.
% ---------------

\maketitle

\begin{abstract}
Motion modelling with block-based architecture has been widely used in video coding where a frame is divided into fixed-sized blocks that are motion compensated independently. This often leads to coding inefficiency as fixed-sized blocks hardly align with the object boundaries. Although hierarchical block-partitioning has been introduced to address this, the increased number of motion vectors limits the benefit. Recently, approximate segmentation of images with cuboidal partitioning has gained popularity. Not only are the variable-sized rectangular segments (cuboids) readily amenable to block-based image/video coding techniques, but they are also capable of aligning well with the object boundaries. This is because cuboidal partitioning is based on a homogeneity constraint, minimising the sum of squared errors (SSE). %Therefore, motion modelling with variable-sized cuboidal blocks is able to exploit semantic correlations by following displacements of approximate object boundaries. Video surveillance applications will benefit as objects can be tracked in real-time by avoiding the high computational burden of accurate segmentation. 
In this paper, we have investigated the potential of cuboids in motion modelling against the fixed-sized blocks used in scalable video coding. Specifically, we have constructed motion-compensated current frame using the cuboidal partitioning information of the anchor frame in a group-of-picture (GOP). The predicted current frame has then been used as the base layer while encoding the current frame as an enhancement layer using the scalable HEVC encoder. Experimental results confirm 6.71\%--10.90\% bitrate savings on 4K video sequences.
	
\end{abstract}


\begin{comment}

\renewcommand\footnoterule{\rule{.96\linewidth}{1 pt}}
\let\thefootnote\relax\footnotetext{This research was partially supported by Australian Research Council (ARC) Discovery Project DP190102574 }

\end{comment}

\section{Introduction}
In recent times, the multimedia technology is developed remarkably. The demand of higher resolution and better quality videos by the users is increasing day-by-day. In addition, due to the cheaper and easier access to higher resolution cameras, a huge amount of video content is being generated every day. Therefore, to store, process and transmit this outpouring video data, 
%The modern digital revolution has caused an uprising of video and video-based applications. Due to higher demands, the data required to store and transmit digital videos increases day-by-day. Therefore, 
highly effective and efficient video coding algorithms are required.  A video sequence is a collection of frames (static images). For a small unit of time, a video sequence captured for a particular scene has a number of frames that exhibit spatiotemporal correlations. In other words, the foreground objects and the background move within one frame to create the subsequent frames in the video sequence. Often the movement of backgrounds in these frames is very little to zero, and the movement of the foreground objects are only visible to a certain extent. Therefore, a huge amount of temporal redundancy exists between the frames of a particular video sequence, and the effectiveness of the intra-frame video compression can be improved by removing the temporal redundancies. 

This is achieved by motion estimation and compensation techniques. Traditionally, it has been performed by %Full search or 
block-matching algorithms where the current frame is divided into a number of fixed-sized macroblocks and then compare each block to its corresponding block and a neighbourhood area in the reference frame to find the best-matched block from the reference frame. This matching is done content-wise. A reference frame can be the immediate past or future to the current frame or a distant neighbour in the temporal order of video frames. A current frame’s block often found its best match in the neighbourhood of the reference frame’s corresponding block. The displacement between the best-matched block and the corresponding block in the reference frame is recorded and called motion vector. In this way, motion vectors are calculated for all macroblocks in the current frame. With the help of these vectors, the current frame can be predicted from the reference, and by adding the residual generated at the encoder, it can be reconstructed at the  decoder. Therefore, instead of transmitting all frames in a sequence, it is only required to transmit the reference frame(s) along with the motion vectors and residual signals associated with the other frames within the same group-of-picture (GOP) \cite{block-review}.

To accommodate the continuously increasing demand of higher resolution videos, the video coding standards have been advanced over the time. Therefore, the corresponding frame portioning structures have also been evolved. Older video coding standards like MPEG-2 \cite{itu1995generic} partitions frames into fixed sized blocks of $ 16 \times 16 $ pixels. The H.264/AVC \cite{wiegand2003overview} video coding standard partitions 
%uses a slightly different block structure where
a frame into macroblocks of $ 16\times16 $ pixels as the maximum size. This macroblocks can be further divided into $ 16 \times 16 $, $ 8 \times 8 $ and $ 4 \times 4 $ pixels for intra prediction and into $ 16 \times 16 $, $ 16 \times 8 $, $ 8 \times 16 $ and $ 8 \times 8 $ for inter-prediction \cite{paul2009efficient, paul2009video}. This structure may be satisfactory for CIF or QCIF video sequences \cite{ma2007high}, but for high-definition (HD) and 4K videos, macroblocks of relatively larger size are required \cite{kwon1997very}. This is due to frame sizes being larger compared to CIF or QCIF formats \cite{kim2012block} and also $ 16\times16 $ blocks may not be able to represent the spatial correlations efficiently in HD frames \cite{articleHEVC}. In contrast, larger block sizes are able to model spatial redundancy more effectively in high-resolution videos and it can also increase the efficiency and flexibility of the encoder \cite{ccetinkaya2021ctu}.

%The H.264/AVC video coding standard partitions frames into $ 16\times16 $ pixels macroblocks \cite{wiegand2003overview}. This size may be satisfactory for CIF or QCIF video sequences \cite{ma2007high}, but for high-definition (HD) videos, macroblocks of relatively larger size are required \cite{kwon1997very}. This is due to image sizes being larger compared to CIF or QCIF formats \cite{kim2012block} and also $ 16\times16 $ blocks may not be able to represent the spatial correlations efficiently in HD frames \cite{articleHEVC}. The H.265/HEVC standard, therefore, introduces larger block sizes where a slice can be partitioned into multiple coding tree units of any size between $ 8\times8 $ and $ 64\times64 $ \cite{sullivan2012overview}. Nevertheless, partitioning blocks with fixed-sized blocks has some limitations. For example, a block in the current frame may contain parts of more than one foreground objects in close proximity. If these two objects move apart in the reference frame, it would be challenging to find a matching block in the corresponding block's neighbourhood area in the reference frame and hence, block-matching performance will be compromised. This happens due to partitioning frames artificially into fixed-sized blocks. This issue can be partially solved by considering a pyramid of variable size blocks. However, it will reduce the efficiency of the algorithm~\cite{ugur2010low}. 

In accordance with the above discussion,  H.265/HEVC \cite{sullivan2012overview} introduced a block partitioning structure named coding tree unit (CTU) that divides frames in variable sizes between $ 8\times8 $ and $ 64\times64 $. With this structure, HEVC can provide a bit rate savings of approximately 50\% for equivalent perceptual quality compared to prior video coding standards, specially for high-resolution videos \cite{ohm2012comparison}. To improve the HEVC performance, many modifications \cite{correa2014fast, badry2020decision} and extensions to HEVC have been proposed in the literature. One such extension is scalable HEVC (SHVC) \cite{boyce2015overview} which provides spatial, signal-to-noise ratio, bit depth, and color gamut scalability in addition to the temporal scalability offered by HEVC. One of the advantages of SHVC is, a video is encoded in more than one layer and provides improved video quality. The lowest quality video at base layer is encoded first and used as a reference to encode one or more enhancement layers. Therefore, in SHVC, to inter-code a current frame, along with the temporally available already coded frame, inter-layer frame is also available as references.

In spite of recent advancements in video coding standards, most of them still uses block partitioning which has some limitations. For example, a block in the current frame may
contain parts of more than one foreground objects in close proximity. If these two objects move apart in the reference frame, it would be challenging to find a matching block in the corresponding block’s neighbourhood area in the reference frame and hence, block-matching performance will be compromised.
This happens due to partitioning frames artificially into fixed-sized blocks. One probable approach to deal with the above issue effectively and efficiently is to partition the current frame based on its content. More specifically, partition the frame (image) into segments that contain entire objects or part of one object only, instead of the artificial fixed-sized blocks that may contain the parts of more than one objects. This goal can be achieved by the Cuboidal Partitioning of Image Data (CuPID) algorithm \cite{murshed2017cuboid, murshed2018enhanced} as it progressively partitions an image into two homogeneous rectangular segments (cuboids). Therefore, the objects or parts (i.e., image contents) that exhibit homogeneity will be partitioned into separate cuboids. It is implausible for a single cuboid to represent more than one objects or parts unless they are content-wise homogeneous. Based on this property, we argue that when an object moves, the movement of corresponding cuboid(s) should be similar, at least for rigid body objects. Therefore, the motion estimation and compensation performed on cuboids will be more accurate than using fixed-sized blocks. Also, the motion modelling with variable-sized cuboids is able to exploit semantic correlations by following displacements of approximate object boundaries.

Although the usage of cuboids will yield better quality reconstructed video frames at the decoder, the bits required to pass the cuboid information of each candidate frame will make the system inefficient and expensive. So, there is a trade-off between effectiveness and efficiency while using the cuboids for motion estimation. However, this issue can be solved because in most cases, GOP frames are from the same scene, and there are little movements among the frames in a GOP. Therefore, cuboids in the current frames will be generated mainly in the same locations as cuboids in the reference frame. Thus, instead of using cuboids of each frame for the corresponding frame's motion estimation, we argue that the cuboids of a reference frame are enough to perform the motion estimation of each frame in the GOP. 

In this paper, our contribution is to investigate the performance of motion estimation and compensation using cuboids i.e., variable-sized blocks rather than the traditional fixed-sized blocks. We argue, as cuboids are better aligned to object boundaries, the residual motion estimation error should be lowered compared to the fixed-sized block-based motion modelling. In addition, instead of obtaining cuboids for each frame, we also investigate the scenario when CuPID algorithm is applied to the anchor frame only, and the resulting cuboidal partitioning is applied to all other frames in the GOP for motion modelling to reduce the bit rate costs and computational time associated to cuboid generation in the individual frame. We argue that by considering this approach, bit transmission rate will be lowered with little to no compromise to the encoding quality. 

The rest of the paper is organised as follows. An overview of the CuPID algorithm is provided in Section II, followed by a discussion on the advantage of cuboids over fixed-sized blocks in motion modelling in Section III. The proposed method is explained in Section IV. Experiment results with setup are provided in Section V. Section VI discusses the subjective quality analysis and finally,  Section VII concludes the paper.

\section {Overview of CuPID}

In this section, we provide an overview of the CuPID segmentation algorithm \cite{murshed2018enhanced}. Instead of accurately segmenting individual objects in an image, an orthogonal linear basis pair (e.g., lines at 0° (horizontal) and 90° (vertical)) is progressively used to hierarchically split image data so that shapes of all objects are approximated jointly. For an image $I$ of $ X \times Y $ pixels, all possible $ X+Y-2 $ splits, $ X-1 $ vertical splits and $ Y-1 $ horizontal splits, are considered to find the best-split  $ s^* $ that minimises the sum of squared errors (SSE) for the feature space $  \mathcal{F} $ in consideration e.g., the RGB or YCbCr colour space. Among the existing cuboids, the one with the best of best-splits is chosen to split next. 

The progressive best-splits can be represented by a binary split-tree where the left child of a node denotes the left (for vertical split) or top (for horizontal split) half and the right child denotes the remaining half. The split-tree of a partitioning with $n$ cuboids has $2n - 1$ nodes, $n - 1$ split-nodes and $n$ leaf-nodes. The type of each node can be encoded with 1 bit. For each split-node of (say) $W \times H$ pixels, the best-split can be encoded with $\log_2{(W + H - 2)}$ bits. Hence, the split-tree $I$ partitioned with $n$ cuboids can be encoded with $O((2n - 1) + (n - 1) \log_2{(X + Y)}) = O(n\log_2{(X + Y)})$ bits.

\section{Advantage of using cuboids over fixed-sized blocks in motion modelling}

\begin{figure}
	\centering
	\subfigure[]{\includegraphics[width=0.22\textwidth]{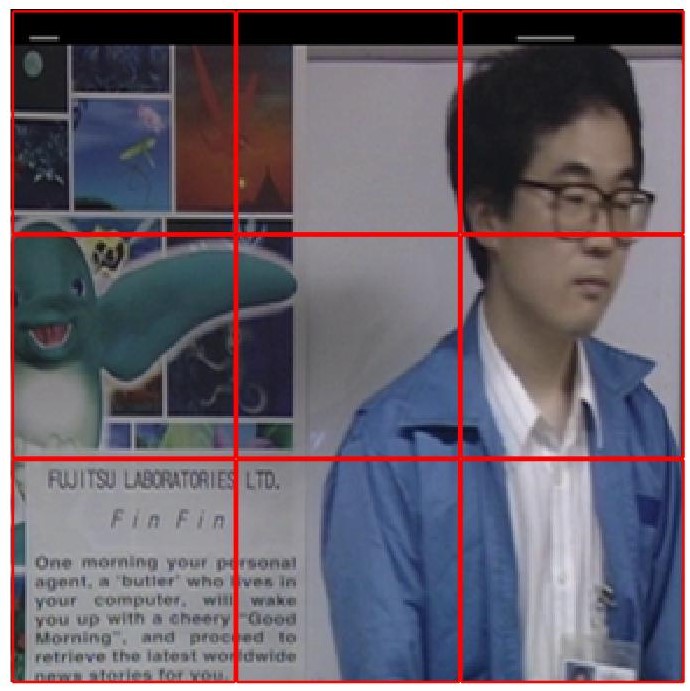}}  
	\subfigure[]{\includegraphics[width=0.22\textwidth]{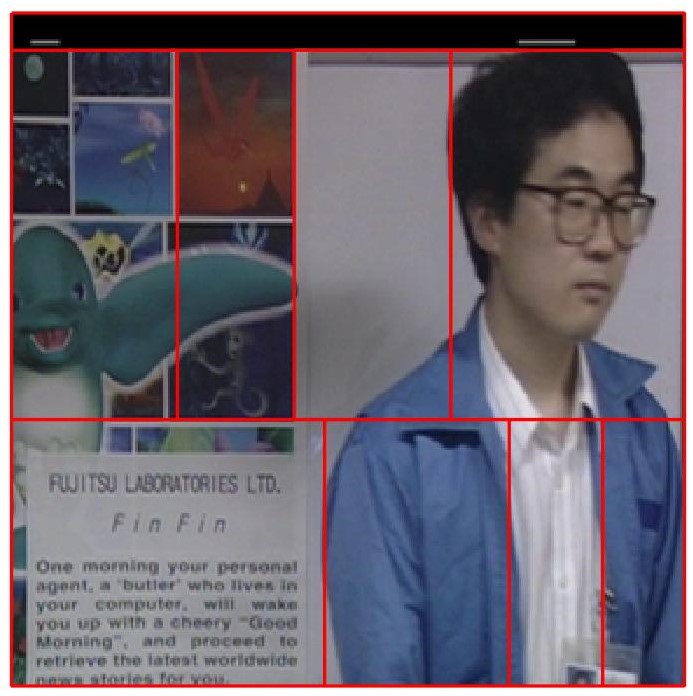}} 

	\caption{(a) Blocks, (b) Cuboids obtained on a sample frame from Bowing sequence.}
	\label{fig:cub_block}
\end{figure}

In this section, the advantage of cuboids over fixed-sized blocks for effective motion modelling is discussed.
CuPID recursively partitions images into homogeneous cuboids. So, that the intra-cuboid contents are homogeneous, and inter-cuboid contents are heterogeneous to each other. This can result in various sizes of cuboids as objects (or parts of it) can exhibit homogeneity in a wider or a smaller image region as shown in Figure \ref{fig:cub_block}(a). In contrast, an image will be divided into fixed-sized blocks as per block-matching methods as shown in Figure \ref{fig:cub_block}(b).

For the sake of simplicity, a CIF video sequence is considered instead of high-resolution sequences. The sample frame in Figure \ref{fig:cub_block} is taken from bowing sequence \cite{bowing}. This sequence begins with a scene of a presentation board, and eventually a man enters into the scene from the right side to the centre, bows down and up, and finally exits the scene from the right side, leaving the initial scene of the presentation board. Therefore, in this sequence, only the man is moving object and motion estimation, compensation of the sequence will be based on this moving object only. 

It can be easily seen from the  Figure \ref{fig:cub_block} that cuboids have captured different parts of the object of interest effectively. The top part (head to the neck-shoulder joint) is represented by only one cuboid. Whereas, the same area of the object, especially the face of the man is divided into two blocks. Similarly, most of the bottom part of the object is precisely represented by three cuboids. Even the shirt and jackets are also separated by different cuboids. In contrast, the same area is represented by three blocks with poor correlation among the contents of each block. Therefore, during the motion search within a GOP, it is highly likely that a cuboid in the current frame will find an almost similar cuboid in the reference frame. In contrast, the current frame's blocks may need to sacrifice a higher degree of content-matching while looking for the best-matched block in the reference frame.

Due to the above advantage of cuboids over the blocks in motion estimation, we have proposed to partition current frames in a GOP using CuPID to obtain cuboids, and thereafter, the motion estimation and compensation have been performed using these cuboids. %However, if we obtain cuboids for each current frame, it will be an inefficient process to transmit this cuboid information for each frame from encoder to decoder side as it will increase the transmission bitrate significantly higher. To overcome this situation, we proposed not to apply CuPID to each current frame in GOP instead we only obtain cuboids of the reference frame and use these cuboids for motion estimation and compensation of  current frames in a GOP. This is a meaningful approach as object(s) motion in GOP is minimal. Therefore, reference frame cuboids representing objects (or parts of objects) are mostly valid to represent the corresponding objects in the successive candidate frames.    

\section{Proposed Method}

The strength of CuPID is the way it partitions frames. Cuboid information have been utilised in \cite{9557821, 9287138} to obtain a coarse version of I-frames and current frames. Coarse frames have been created by replacing the pixels values in a cuboid by the mean value of its content. Although by using coarse frames at the base-layer of scalable HEVC setting has shown  improvement in coding gain, no motion-related information is included here.

CuPID uses object-oriented partitioning. Therefore, it can capture the shape and boundaries of the object more effectively compared to fixed-sized block-partitioning. To explore the strength of CuPID in video coding, we have used cuboids in motion modelling. Specifically, we have partitioned each current frame in a GOP by CuPID. These cuboids obtained from each current frame are then used for motion estimation and compensation with respect to a reference frame. The motion compensated current frame is then used as an additional reference frame at the base layer of scalable HEVC encoder to encode the current frame. The proposed method is explained from the point of view of a scalable video coding solution, in particular, in the form of quality scalability/SNR scalability. Hence, for experimental analysis, the SHM reference software \cite{shm} for scalable HEVC is used. For both the anchor and proposed cases, a base layer has been provided.  In the proposed case, the base layer is the predicted frame generated by the proposed cuboid-based approach. As for the anchor, a lower quality version of the current frame is supplied as the base layer to ensure that the quality (PSNR) of this base layer matches closely with the quality of the base layer used in the proposed approach.

\label{sec:motion-modelling}
In a GOP, it is assumed that scene change does not occur significantly. In addition, the background remains the same, and the foreground objects exhibit little to no motion in most cases. Based on this understanding, cuboids obtained on a single frame in a GOP can still represent the object shape and boundaries effectively. Therefore, it is not required to apply CuPID on individual current frames. Instead, we have obtained the cuboids only from the I-frame in a GOP and used these information for the motion modelling of the rest of the P frames. This is a meaningful approach as object(s) motion in GOP is minimal. Therefore, I-frame cuboids representing objects (or parts of objects) are mostly valid to represent the corresponding objects in the successive P-frames.

\begin{figure*}
	\centering
	\subfigure[]{\includegraphics[width=0.49\textwidth]{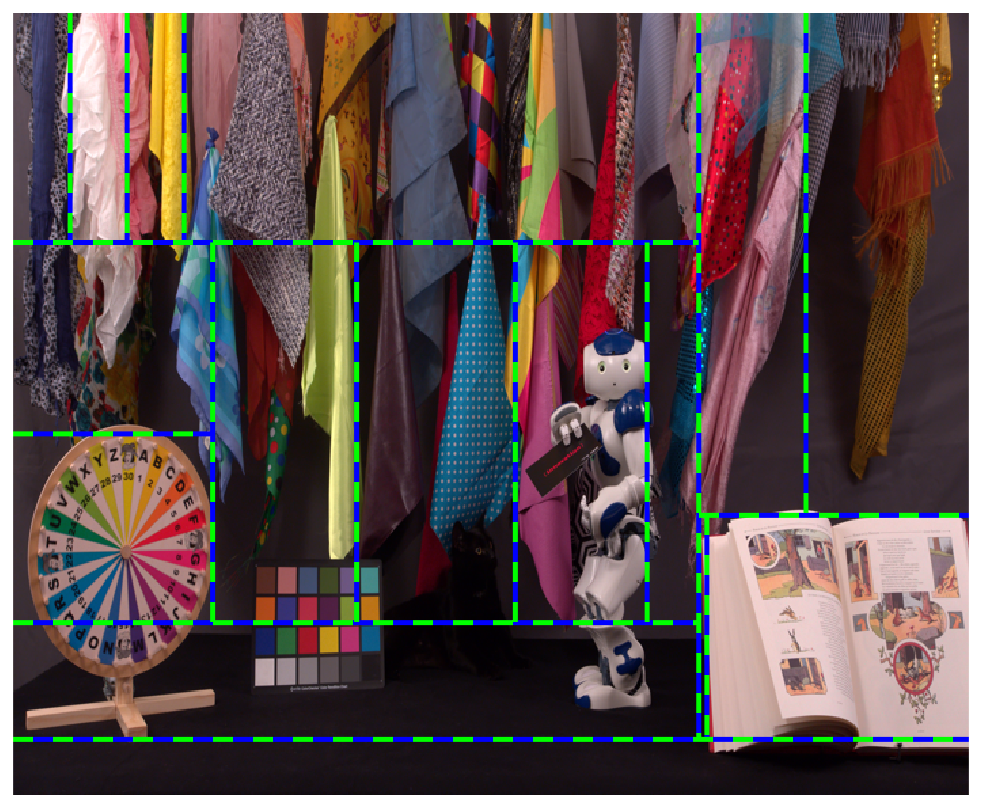}}  
	\subfigure[]{\includegraphics[width=0.49\textwidth]{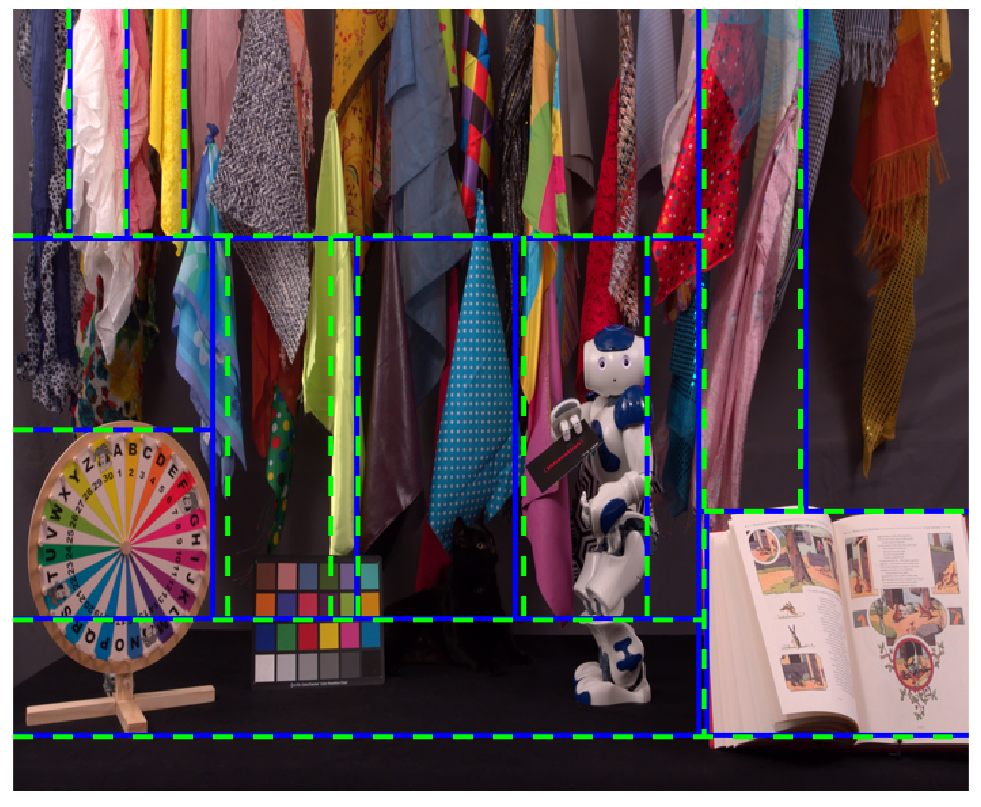}}
	
	\caption{Cuboid partition lines on (a) Frame 2, (b) Frame 10 of first GOP of CatRobot sequence. Solid blue  lines: Frame 1 partition lines. Dashed green lines: current frame partition lines.}
	\label{fig:catRobot_proposal}
\end{figure*}

By adopting this approach, motion estimation performance may get degraded to a certain extent, especially for the P frames which are further from the I-frame. However, there is an advantage in terms of bit signalling. As only one frame's cuboid information needs to be transmitted, instead of cuboids of each P frames in a GOP, there is a saving in terms of bit transmission. Therefore, eventually, coding gain will be improved. This proposal is supported by an example using Figures. \ref{fig:catRobot_proposal}(a) and \ref{fig:catRobot_proposal}(b) belonging to the CatRobot sequence. Frame 1 of the first GOP of this sequence is an I-frame and Frame 2 and Frame 10 from the same GOP are  current frames. At first, CuPID is applied on I-frame and obtained cuboid information is placed on both current frames using solid blue lines. Similarly, CuPID is applied on this current frames independently and obtained cuboid information are placed on the corresponding frames, respectively using dashed green lines. For the ease of representation, the number of cuboids considered is $ 16 $ for this example. From Fig. \ref{fig:catRobot_proposal}(a), it is clear that I-frame and Frame 2 result in the same partition lines as they overlap. I-frame and Frame 10, as shown in Fig. \ref{fig:catRobot_proposal}(b) results in the same partition lines mostly. Only two pairs of lines did not overlap, but they are co-locating in close proximity.
%using Figures \ref{fig:catRobot_proposal}(a) and \ref{fig:catRobot_proposal}(b) belonging to the CatRobot sequence. Say, the Frame 1 of the first GOP of this sequence is considered an I-frame and Frame 2 and Frame 10 from the same GOP are considered P-frames. At first, CuPID is applied on I-frame and obtained cuboid information is placed on Frames 2 and 10 using solid blue  lines. Similarly, CuPID is applied on Frames 2 and 10, and obtained cuboid information are placed on the corresponding frames, respectively using dashed green lines. For the ease of representation, the number of cuboids considered is $ 16 $ for this example. However, this number is much higher while performing experiments, as explained in \hyperref[sec:cuboid_number]{Section 5}. From Figure \ref{fig:catRobot_proposal}(a), it is clear that I-frame and Frame 2 result in the same partition lines as they overlap. I-frame and Frame 10, as shown in Figure \ref{fig:catRobot_proposal}(b) results in the same partition lines mostly. Only two pairs of lines did not overlap, but they are in close proximity. 
Therefore, we can argue that only I-frame's cuboid information is sufficient for the motion modelling of the P-frames in the corresponding GOP.

\begin{table}
	\centering
	\caption{Total bitrate calculation for the comparing methods }
	\label{bitrate}
	\begin{tabular}{|>{\centering}m{.6in}|  >{\arraybackslash}m{2.4in} |}
		\hline
		Method &  \; \; \; \; \; \; \; \; \; \; Total bitrate    \\ \hline
		Scalable HEVC & Summation of bitrates from base and enhancement layers.
		\\ \hline
		Coarse rep. \cite{9557821} & Summation of enhancement layer bitrate and the bitrate required to transmit coarse frames.
		\\ \hline
		Proposed & Summation of enhancement layer bitrate and the bitrate required to transmit motion-compensated frames.
		\\ \hline
		
	\end{tabular}
\end{table}

\begin{table*}
	\centering
	\caption{Bj{\o}ntegaard delta gains obtained over the proposed method compared to Scalable HEVC and Coarse rep. \cite{9557821} }
	\begin{tabular}{|>{\centering}m{2cm}| >{\arraybackslash}p{2cm}p{2cm}|>{\centering}p{2cm}  p{2cm}|}
		\hline
		\multirow{ 2}{*}{Sequence}   & \multicolumn{2}{c|}{Compared with Scalable HEVC} &\multicolumn{2}{c|}{Compared with Coarse rep. \cite{9557821}}  \\

		& Delta rate  & Delta PSNR& Delta rate& Delta PSNR \\ \hline
		Marathon & -7.15 \%  & \;  +0.18 dB & -6.44\% &  \; +0.13 dB \\ \hline
		Campfire & -7.45 \%  & \;  +0.22 dB & -6.12\% &  \; +0.15 dB \\ \hline
		CatRobot& -10.90 \%  & \;  +0.14 dB & -5.28\% &  \; +0.07 dB \\ \hline
		DaylightRoad2& -6.71 \%  & \;  +0.21 dB & -3.10\% &  \; +0.12 dB \\ \hline
	\end{tabular}
	
	\label{table2}
\end{table*}

\begin{comment}

\begin{figure*}
	\centering
	\subfigure[]{\includegraphics[width=0.24\textwidth]{Marathon-threeComp-new.png}}  
	\subfigure[]{\includegraphics[width=0.24\textwidth]{campfire-threeComp-new.png}} 
	\subfigure[]{\includegraphics[width=0.24\textwidth]{CatRobot-threeComp-new.png}} 
	\subfigure[]{\includegraphics[width=0.24\textwidth]{DaylightRoad2-threeComp-new.png}} 
	
	\caption{Comparison of Rate-Distortion performances over (a) Marathon, (b) Campfire, (c) CatRobot, (d) DaylightRoad2. }
	\label{fig:r-d-graphs}
\end{figure*}

\end{comment}

\begin{figure}
	\centering
	\includegraphics[width=.48\textwidth]{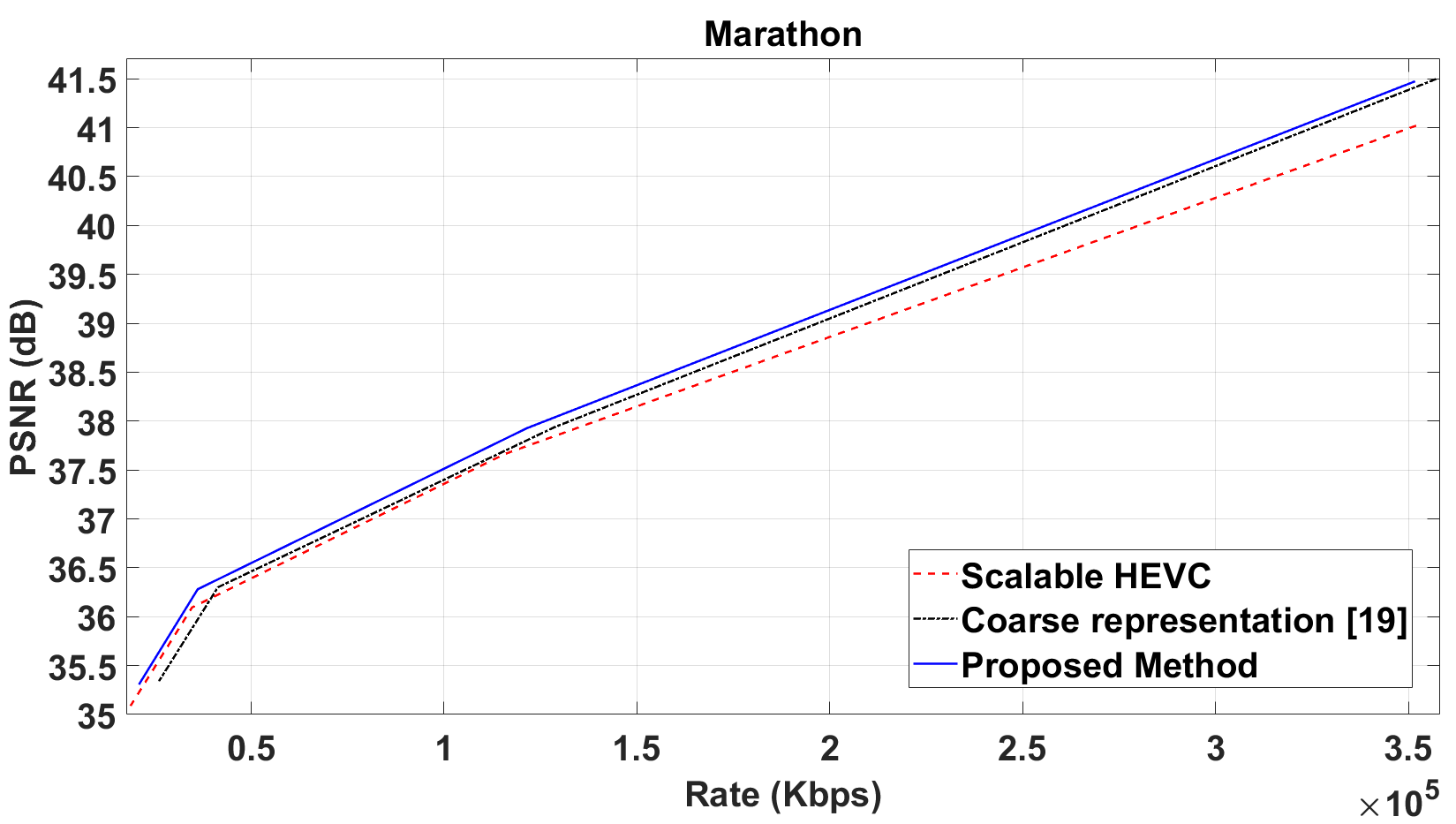}
	\caption{Comparison of Rate Distortion performances over Marathon sequence}
	\label{fig:marathon-r-d}
\end{figure}

\begin{figure}
	\centering
	\includegraphics[width=.48\textwidth]{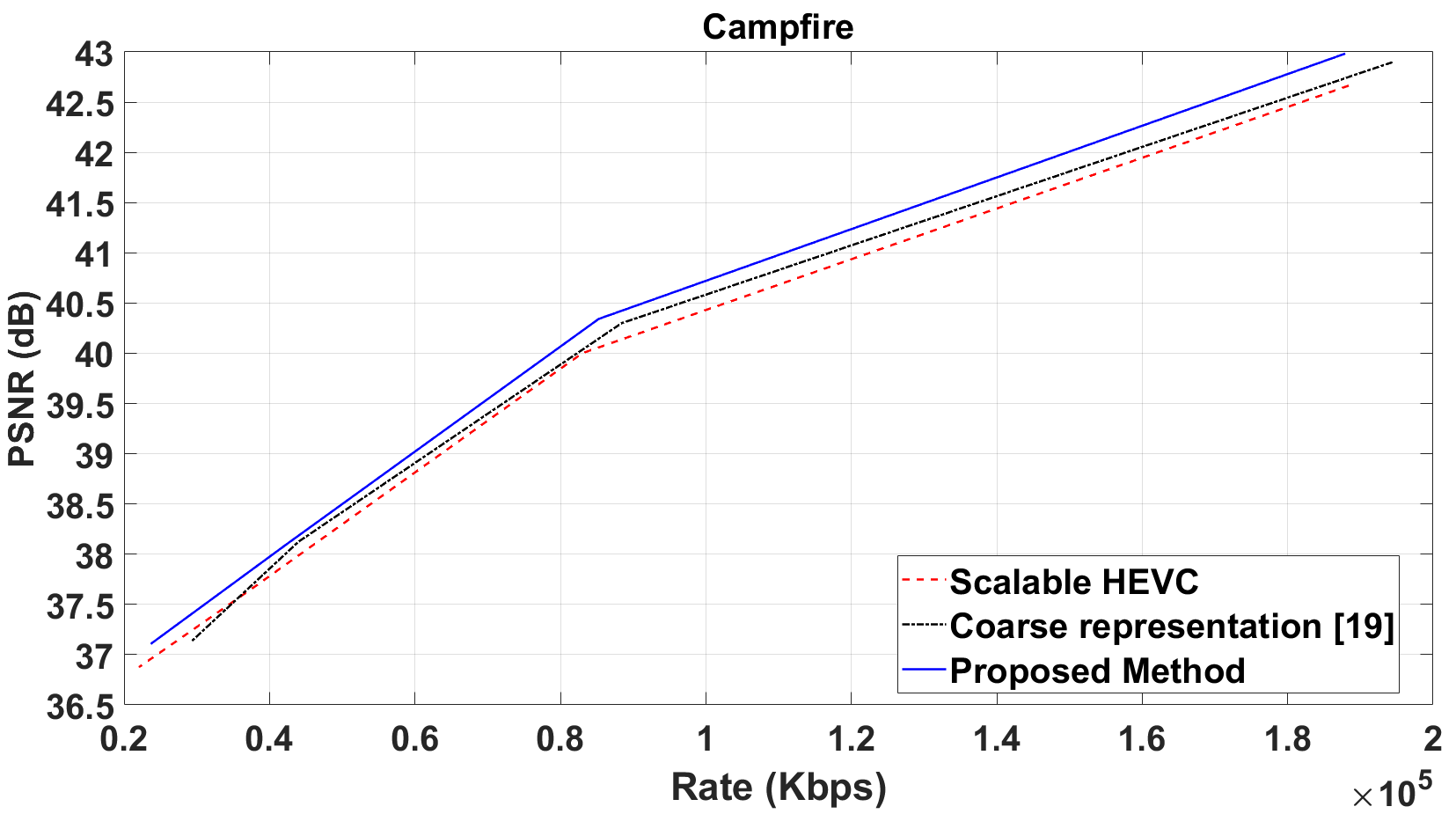}
	\caption{Comparison of Rate Distortion performances over Campfire sequence}
	\label{fig:campfire-r-d}
\end{figure}

\begin{figure}
	\centering
	\includegraphics[width=.48\textwidth]{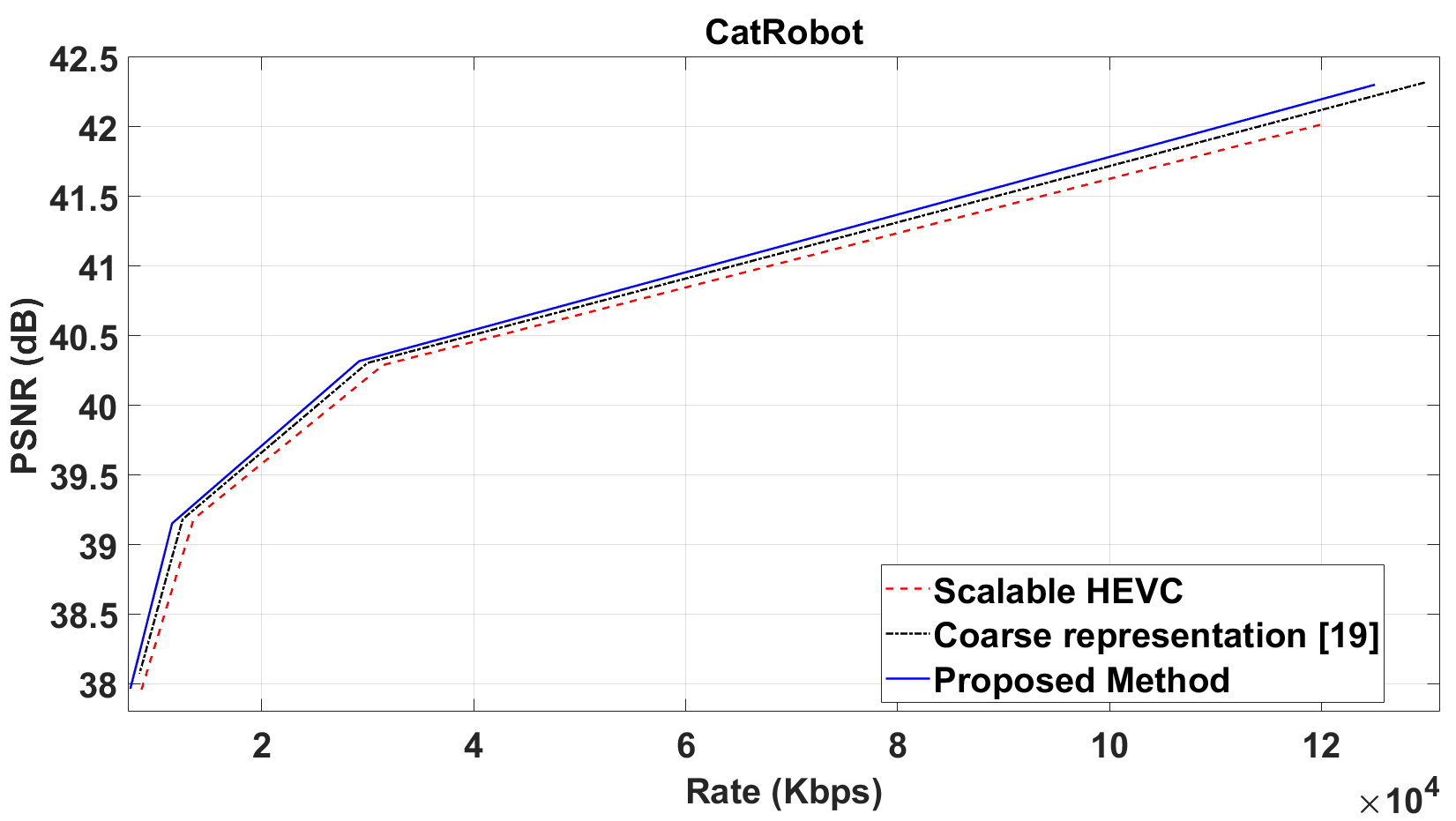}
	\caption{Comparison of Rate Distortion performances over CatRobot sequence}
	\label{fig:catrobot-r-d}
\end{figure}

\begin{figure}
	\centering
	\includegraphics[width=.48\textwidth]{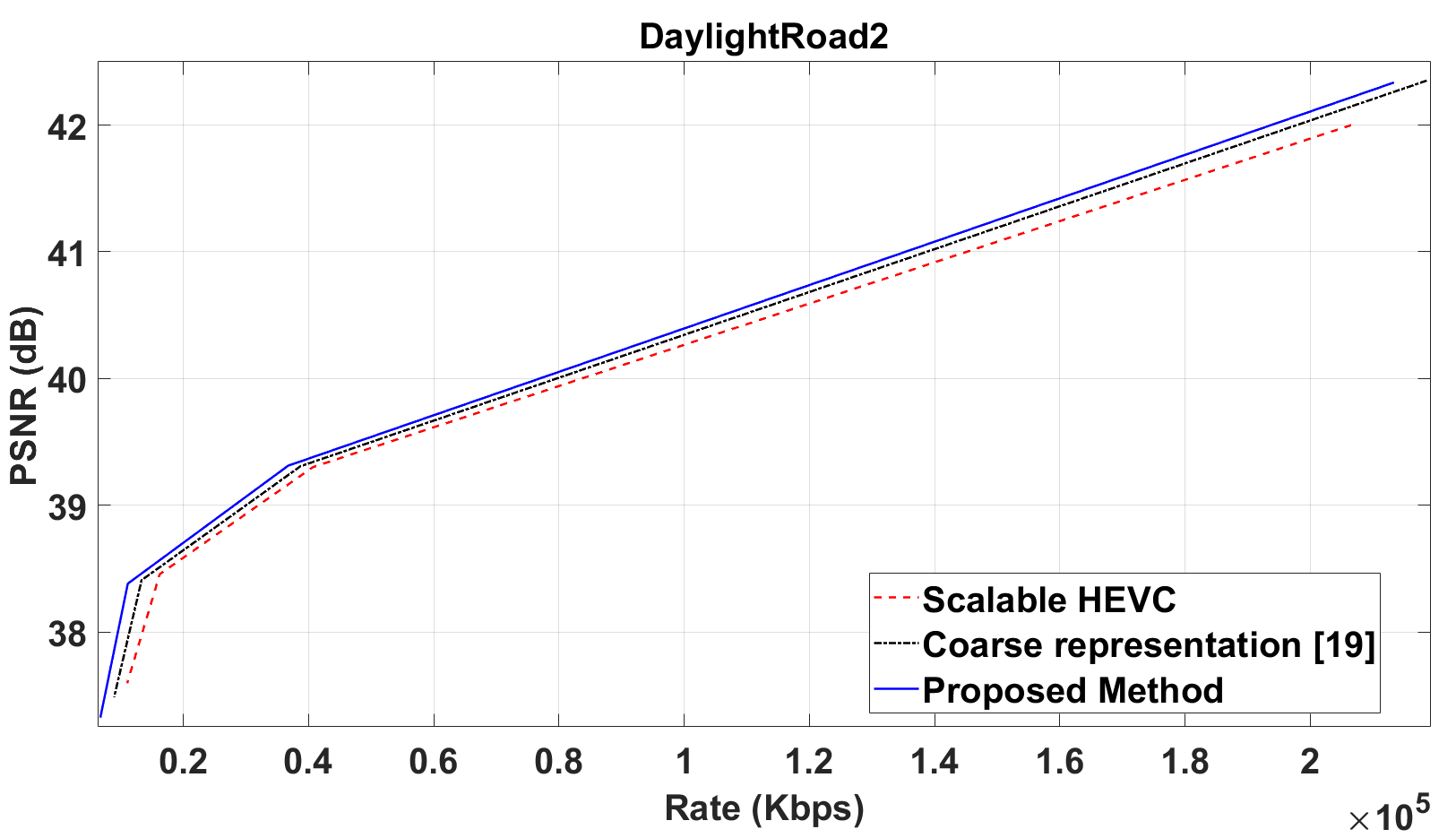}
	\caption{Comparison of Rate Distortion performances over DaylightRoad2 sequence}
	\label{fig:daylight-r-d}
\end{figure}

\begin{figure}
	\centering
	\includegraphics[width=.48\textwidth]{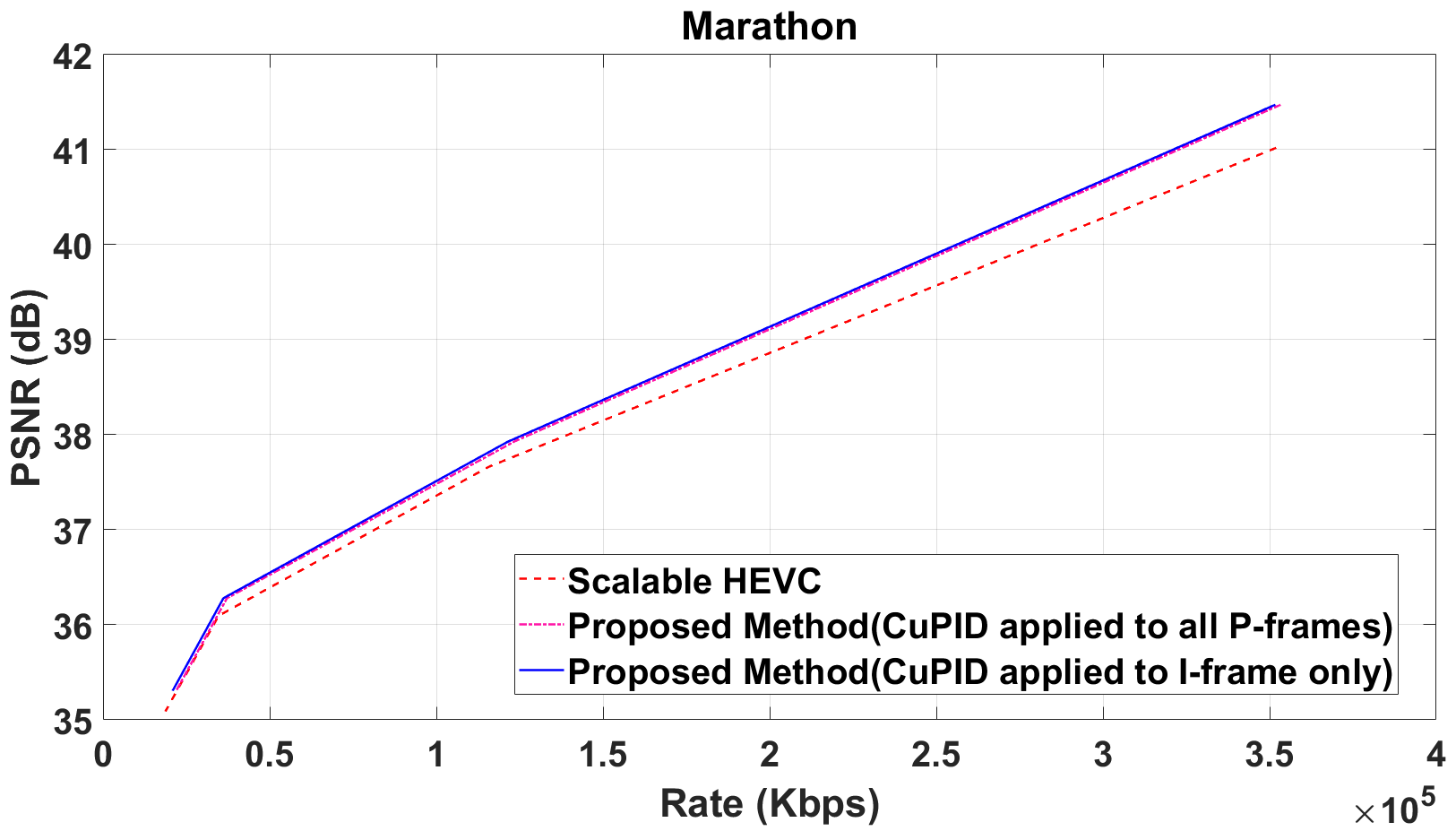}
	\caption{Comparison of Rate Distortion performances over Proposed methods with Scalable HEVC.}
	\label{fig:marathon-proposedComp}
\end{figure}

\section{Experiment and Results}
In this section, experiment settings and results are discussed. We have considered four 4K video sequences to investigate the performance of our proposed method.  The sequences are Marathon, Campfire, CatRobot, DaylightRoad2. The first two are selected from \cite{6603201} and the last three are part of JVET class A1 and A2 test sequences \cite{jvet-2018}.

In a traditional video coding system, the first frame in a GOP is intra coded as an I-frame. The consecutive frames in the GOP are inter-coded as P-frames using already coded frames. In our proposed approach, we have modified this architecture to provide the encoder with an additional path to consider the object shape and boundaries while coding each frame. For the additional path, the first frame in the GOP is encoded by the scalable HEVC encoder \cite{shm}. The resultant bitstream is decoded with the base layer information only. This decoded frame is referred to as I-frame and  used as a reference to obtain the next motion-compensated P-frame using cuboid information as explained in \hyperref[sec:motion-modelling]{Section 4}. Similarly, the consecutive P-frames are predicted using the previously predicted frames. These motion-compensated frames are fed to the base layer of scalable HEVC (SHM 12.4) software \cite{shm} to provide an additional path to code P-frames. While predicting a unit during the encoding of frames, the encoder  optimally decides based on the rate-distortion (RD) how to utilise the information coming from the additional path which contains motion modelling based on the object shape and boundaries.

Motion estimation and compensation has been performed similar way as it is traditionally performed for block-based motion modelling \cite{full-search}, but with cuboids instead of fixed-sized blocks. The number of cuboids obtained by the application of CuPID is provided by the user. In this paper, the number of cuboids parameter is determined by the following approach: If a frame is partitioned by the fixed-sized blocks, then the number of blocks generated is the number passed to the CuPID algorithm. For example, the resolution of frames belonging to the CatRobot sequence is $2160\times3840\times3$. If the size of fixed-blocks are considered as $ 32\times32 $, the total number of partitions is $ 8040 $ which is provided to CuPID algorithm to generate $ 8040 $ cuboids on CatRobot frames. We have also preliminarily observed the impact of different cuboid numbers on the performance of the proposed method and the detailed analysis will be provided in a future work.

\label{sec:cuboid_number}

The RD performance of the video sequences are obtained using four different quantization parameter (QP) values: 22, 27, 32, 37. These QP values belong to the enhancement layer and in each case, the base layer QP value is considered two  more than the corresponding enhancement layer QP. The RD performance of our proposed method is compared with the scalable HEVC, and the method based on coarse representation \cite{9557821}. To obtain the results on scalable HEVC, original uncoded sequences are fed to both enhancement and base layers. For our proposed method and the coarse representation method, enhancement layer accepts the original sequence but, the base layer is provided with motion-compensated frames and the coarse frames, respectively.

To obtain the RD performances, the total bitrate calculation for the three comparing methods is provided in Table \ref{bitrate}. The PSNR values are obtained from the enhancement layer. The RD graphs of the sequences under consideration are provided in Figure \ref{fig:marathon-r-d}, \ref{fig:campfire-r-d}, \ref{fig:catrobot-r-d}, \ref{fig:daylight-r-d}  respectively for Marathon, Campfire, CatRobot, DaylightRoad2. From the graphs, it is evident that the performance of our proposed method is higher than the two other compared methods. It can also be observed from the RD graphs that the average performance of the proposed method is higher in lower QPs or at higher bitrate. This is because, at a higher bitrate, a lot of bits is allocated to preserve the detail of foreground objects like shape and boundaries.

In our proposed method, we argue that in a GOP, only I-frame's cuboid information is enough for modelling the motion in all P-frames instead of obtaining cuboids for each P-frame. To establish our argument, we have presented a RD graph in Figure \ref{fig:marathon-proposedComp} on Marathon sequence. From the graph, it is clear that obtaining cuboids for each P-frame and using it for motion modelling results in better performance than scalable HEVC. However, its performance is slightly lower than the scenario when motion modelling is done only with I-frame's cuboid information. Although, obtaining cuboid for each P-frame can increase the motion estimation and compensation quality, the bits required to transmit these cuboid information degrades the overall performance compared to the scenario where only I-frame's cuboid information is transmitted. Based on this observation, for the rest of the sequences considered, we have shown the results of the proposed method only with the scenario when I-frame's cuboids are used for motion modelling in a GOP.

To compare the effectiveness of our proposed method with the scalable HEVC and \cite{9557821}, we have also provided the Bj{\o}ntegaard deltas\cite{bjontegaard2001calculation} in Table \ref{table2}. The proposed method results in bitrate savings for all sequences against the two comparing methods. As expected, the performance of the scalable HEVC is the least followed by the coarse representation method in terms of bitrate savings as well as PSNR gain. This is because, block partitioning in HEVC cannot capture object shape and boundaries. Whereas, Coarse representation method captures the overall object shape and boundaries using mean pixel values. However, it still lacks motion information. In contrast, our proposed method captures the motion information as well.

The complexity of the proposed method is competitive  with the two compared methods in this paper.  To perform the motion modelling of the current frames which are used as the additional reference at the base layer, the number of blocks and the number of cuboids in case of scalable HEVC and the proposed method, respectively are same. Therefore, the complexity associated with the motion modelling is same for the both methods. In the proposed method, cuboids are obtained by applying CuPID to the I-frame in a GOP. However, the computation complexity of CuPID is minimal \cite{murshed2017cuboid}. Therefore, the overall complexity of scalable HEVC and the proposed method are comparable. Although there is no motion modelling involved in coarse representation method \cite{9557821}, CuPID has been applied to each frame in a GOP and the added complexity is not negligible like when CuPID was only applied to a single frame in a GOP. We have observed that the computing time required to apply CuPID to all current frames is similar to the computing time required for motion modelling of current frames. Therefore, the complexity of \cite{9557821} is also comparable with the proposed method.

\section{Subjective Quality Analysis}

\begin{figure}
	\centering
	\subfigure[]{\includegraphics[width=0.109107\textwidth]{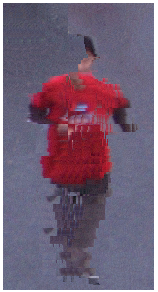}}
	\subfigure[]{\includegraphics[width=0.108\textwidth]{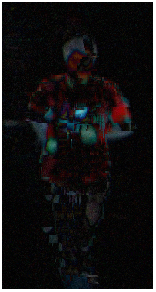}}  
	\subfigure[]{\includegraphics[width=0.113\textwidth]{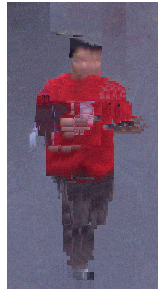}} 
	\subfigure[]{\includegraphics[width=0.108\textwidth]{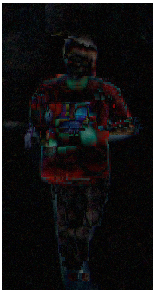}} 
	
	\caption{Subjective quality of a segment in Frame=12, GOP=1 of Marathon sequence: (a)-(b) motion compensated frame (PSNR 21.11 dB) and residual (PSNR 42.98 dB of coded residual) for fixed-sized blocks, respectively, and (c)-(d) the same (PSNRs 22.40 dB and 42.99 dB) for the proposed method.}
	\label{fig:pred_cub_block}
\end{figure}

In this section, we analyse and compare the quality of motion compensated/predicted frames generated using the proposed method and the fixed-sized block-based approach with the help of an example provided in Figure \ref{fig:pred_cub_block} which contains segments of predicted frames (Frame=12, GOP=1) from marathon sequence.  We argue that cuboids can capture shape and boundaries of objects better than the fixed-sized blocks. Therefore, motion modelling using cuboids is more effective than using fixed-sized blocks. This argument is validated by the example in Figure \ref{fig:pred_cub_block}(c) which is a segment of the predicted frame obtained using cuboids and we can see that most parts of the object ( a runner) are better preserved (i.e., predicted from the reference frame) than the segment of the predicted frame obtained using fixed-sized blocks shown in Figure \ref{fig:pred_cub_block}(a). In addition, we have also observed that the quality of the predicted frame (i.e., the frame provided at the base layer before residual coding) belonging to Figure \ref{fig:pred_cub_block}(c) is 6.08\% higher in terms of PSNR than the predicted frame belonging to Figure \ref{fig:pred_cub_block}(a). Therefore, the quality of the residual coded frame belonging to the proposed method is higher than that belonging to the method with fixed-sized blocks.

\section{Conclusion}

In this paper, we have investigated the potential of cuboidal partitioning in motion modelling for scalable video coding. Experimental results show that the performance of our proposed method is superior to two state-of-the-art scalable video coders. In future, we aim to investigate how the number of cuboids impacts the motion modelling performance. At present, the number of cuboids parameter is user-defined. In future, we also aim to propose a method that will automatically determine the optimal number of cuboids.

\bibliographystyle{elsarticle-num}
\bibliography{videoref}	
\balance

\end{document}